\documentclass[11pt]{article}
\usepackage{fullpage}
\usepackage{times}
\usepackage{url}
\usepackage{latexsym}
\usepackage{tabularx}
\usepackage{array}
\usepackage{epsfig}
\usepackage{amssymb}
\usepackage{multirow}
\usepackage{natbib}
\usepackage{natbibspacing}

\bibpunct{(}{)}{;}{a}{,}{,}

\usepackage{color}

\newcolumntype{x}[1]
{>{\raggedright}p{#1}}
\newcommand{\tn}{\tabularnewline}

\title{An Empirical Comparison of Parsing Methods \\ for Stanford Dependencies}
\author{Lingpeng Kong \quad Noah A.~Smith \\ Language Technologies
  Institute \\ Carnegie Mellon University \\ Pittsburgh, PA 15213, USA
  \\ \texttt{\{lingpenk,nasmith\}@cs.cmu.edu}}

\date{}

\begin{document}
\maketitle
\begin{abstract}
Stanford typed dependencies are a widely desired representation of
natural language sentences,
but parsing is one of the major computational bottlenecks in text
analysis systems.  In light of the evolving definition of the Stanford
dependencies and developments in statistical dependency parsing
algorithms, this paper revisits the question of
\citet{cer2010parsing}:  what is the tradeoff between accuracy and
speed in obtaining Stanford dependencies in particular?  We also explore the 
effects of input representations on this tradeoff:  part-of-speech tags, the novel
use of an alternative dependency representation as input, and
distributional representaions of words.  We find that direct
dependency parsing is a more viable solution than it was found to be
in the past.  An accompanying software release can be found at: \\ \url{http://www.ark.cs.cmu.edu/TBSD}
\end{abstract}
\section{Introduction}
\label{se:intro}

The Stanford typed dependency (SD) representations of grammatical relations,
introduced by \citet{de2008stanford}, have
become a popular text analysis scheme for a variety of NLP
applications such as event extraction \citep{bjorne2009extracting}, error
correction \citep{tetreault2010using}, and machine translation
\citep{stein2010cocktail}.  Relative to other dependency representations, such
as those of \nocite{yamada2003statistical} Yamada and Matsumoto (YM;
2003), SD parses emphasize semantic relations (e.g., relative clauses
are rooted in the verb rather than the complementizer, prepositional phrases in the nominal rather
than the preposition). This may contribute to their attractiveness in
downstream applications \citep{elming2013down}, but it also makes SD parsing more challenging
than YM parsing.  

Broadly speaking, there are two kinds of approaches to obtaining
dependencies.  One, which we call \textbf{c-parsing}, applies
phrase-structure parsing algorithms to obtain constituents, then
extracts dependencies by applying expert-crafted head rules and
perhaps other transformations.  This is the dominant approach for SD
parsing; indeed, the rules\footnote{\url{http://nlp.stanford.edu/software/corenlp.shtml}} are considered
  definitive for the representation, and they are updated frequently.

The second approach, which we call \textbf{d-parsing}, applies
dependency parsing algorithms, directly tackling the grammatical
relations without representing constituents.  These parsers tend to be
faster, and for YM dependencies they achieve the best performance:
\citet{martins2013turning} report 93.1\% unlabeled attachment score on
PTB \S 23, while a state-of-the-art phrase-structure parser
\citep{charniak2005coarse,mcclosky2006effective} achieves 92.5\%.
Several recent dependency parsers \citep{rush2012vine,choi2013transition,zhang2013online} further improve the speed of this approach while preserving accuracy.

The main contribution of this paper is an empirical comparison of a
wide range of
different ways to obtain SD parses.  It follows on an
important study by \citet{cer2010parsing}, which found a 6.9\%
absolute unlabeled (8\% absolute labeled) point gap in $F_1$ between
c-parsing with the best available
phrase-structure  parser and d-parsing with the best available
dependency parser in SD parsing for \textsc{CCprocessed} dependencies
(the most linguistically rich representation).
 \citet{cetinoglu2010lfg} explored similar
strategies for parsing into lexical-functional grammar representations
without phrase-structure parsing.

Since those studies, dependency parsing has advanced, and the definition
of SDs has evolved considerably, so it is worth revisiting the
viability of d-parsing for Stanford typed dependencies.  
For Chinese, \citet{che2012comparison} found greater
success with a second-order dependency parser, Mate
\citep{bohnet2010very}.

This paper's contributions are:
\begin{itemize}
\item We quantify the current tradeoff between accuracy and speed in
  SD parsing, notably
closing the gap between c-parsing and d-parsing to 1.8\%
  absolute unlabeled (2.0\%
  absolute labeled) $F_1$ points
  (\S\ref{se:tradeoff}) for \textsc{CCprocessed} SD parsing.  The current
  gap is 30\% (25\%) the size of the one found by
  \citet{cer2010parsing}.
An arc-factored d-parser is shown to perform a bit better than the
Stanford CoreNLP pipeline, at twenty times the speed.

\item We quantify the effect of part-of-speech tagging on SD parsing
  performance, isolating POS errors as a major cause of that gap (\S\ref{se:pos}).
\item We demonstrate the usefulness of the YM representation as a
  source of information for SD parsing, in a stacking framework
  (\S\ref{se:stacking}).
\item Noting recently attested benefits of distributional word
  representations in parsing \citep{koo2008simple},
  we find that d-parsing augmented with Brown cluster features
  performs similarly to c-parsing with the Stanford recursive neural
  network parser \citep{socher2013parsing}, at three times the speed.
\end{itemize}

\section{Background and Methods}

A Stanford dependency graph consists of a set of ordered dependency
tuples $\langle T, P, C\rangle$, where $T$ is the type of the
dependency and $P$ and $C$ are parent and child word tokens,
respectively.  These graphs were designed to be generated from the
phrase-structure tree of a sentence \citep{de2006generating}. This
transformation happens in several stages. First, head rules are used
to extract parent-child pairs from a phrase-structure parse.  Second,
each dependency is labeled with a grammatical relation type, using the
most specific matching pattern from an expert-crafted set.

There are several SD conventions.  The simplest, \textsc{Basic} SD
graphs, are always trees.  Additional rules can be applied to a
phrase-structure tree to identify
\textsc{Extra} depenencies (e.g., \emph{ref} arcs attaching a relativizer like
\emph{which} to the head of the NP modified by a relative clause), and
then to collapse dependencies involving
transitions and propagate conjunct dependencies, giving the richest
convention, \textsc{CCprocessed}.  In this paper we measure
performance first on
\textsc{Basic} dependencies; in \S\ref{se:ccprocessed} we show that
the quality of \textsc{CCprocessed} dependencies tends to improve as
\textsc{Basic} dependencies improve.

The procedures for c-parsing and d-parsing are well-established \citep{cer2010parsing}; we
briefly review them.    In c-parsing, a phrase-structure parser is
applied, after which the Stanford CoreNLP rules are applied to obtain
the SD graph.  In this work, we use the latest version available at
this writing, which is  version 3.3.0.  In d-parsing, a statistical dependency
parsing model is applied to the
sentence; these models are trained on Penn Treebank trees (\S 02--21)
transformed into \textsc{Basic} dependency trees using the Stanford
rules.  To obtain \textsc{CCprocessed} graphs, \textsc{Extra}
dependencies must be added using rules, then the
collapsing and propagation transformations
must be applied. 

One important change in the Stanford dependencies since
\citet{cer2010parsing} conducted their study is the introduction of
rules to infer \textsc{Extra} dependencies from the phrase-structure
tree.  
(Cer et al.~used version 1.6.2; we
use 3.3.0.) 
We found that, given \emph{perfect} \textsc{Basic}
dependencies (but no phrase-structure tree), the inability to apply
such inference rules accounts for a 0.6\% absolute gap in unlabeled
$F_1$ (0.5\% labeled)
between c-parsing
and d-parsing for \textsc{CCprocessed} dependencies (version 1.6.2).\footnote{In version 3.3.0, 
inference rules have been added to the Stanford CoreNLP package to
convert from \textsc{Basic} to \textsc{CCprocessed} without
a phrase-structure tree. Given perfect
\textsc{Basic} dependencies, 
 there is still a 0.2\% unlabeled (0.3\% labeled) gap
in $F_1$ in PTB \S 22 
(0.4\% and 0.5\% for \S 23).  We added some new rules to help close
this gap by about 0.1 $F_1$ (unlabeled and labeled), but more can be
done.  The new rules are not fine-tuned to \S 22--23; they are given in Appendix~\ref{app:rules}.}

\section{Current Tradeoffs} \label{se:tradeoff}

We measure the performance of different c-parsing and d-parsing
methods in terms of unlabeled and labeled attachment score (UAS and
LAS, respectively) on Penn Treebank \S 22 and \S 23.  We report
parsing speeds on a Lenovo ThinkCentre desktop computer with Core
i7-3770 3.4GHz 8M cache CPU and 32GB memory.  All parsers were trained
using Penn Treebank \S 02--21.  We target version 3.3.0 of SDs
(released November 12, 2013), and, where Stanford CoreNLP components
are used, they are the same version. 

We consider three c-parsing methods:
\begin{enumerate}
\item The Stanford ``englishPCFG'' parser, version 3.3.0
  \citep{klein2003accurate}, which we believe is the most widely used
 pipeline for SD parsing. This model uses additional non-WSJ
training data for their English parsing model.\footnote{See the Stanford Parser 
FAQ at \url{http://nlp.stanford.edu/software/parser-faq.shtml}.}

\item The Stanford ``RNN'' parser, version 3.3.0 \citep{socher2013parsing},
which combines PCFGs with a syntactically untied recursive neural network
that learns syntactic/semantic compositional vector representations. 
Note this model uses distributional representations from external corpus;
see section \ref{se:idr}.

\item The Berkeley ``Aug10(eng\_sm6.gr)'' parser, version 1.7
  \citep{petrov2006learning}.
\item Charniak and Johnson's ``June06(CJ)'' parser
  \citep{charniak2005coarse,mcclosky2006effective}.
  Note this is the self-trained model which uses 2 million unlabeled sentences
  from the North American News Text corpus,  NANC
  \citep{graff1995north}. It is therefore technically semi-supervised.
\end{enumerate}
Each of these parsers performs its own POS tagging.
Runtime measurements for these parsers include POS tagging and also conversion
to SD graphs.

We consider eight d-parsing methods:
\begin{enumerate}\setcounter{enumi}{3}
\item MaltParser liblinear stackproj \citep{nivre2006maltparser} a transition-based
dependency parser that uses the Stack-Projective algorithm.
The transitions are essentially the same as in the
``arc-standard'' version of Nivre's algorithm and 
produce only projective dependency trees \citep{nivre2009non,nivre2009improved}.
In learning, it uses the LIBLINEAR package implemented by \citet{fan2008liblinear}.
This is the same setting as the most popular pre-trained model provided by MaltParser.
\item MaltParser libsvm arc-eager \citep{nivre2006maltparser}, a transition-based
dependency parser that uses the ``arc-eager'' algorithm
\citep{nivre2004incrementality}.  In learning, it
uses LIBSVM implemented by \citet{chang2011libsvm}. This is the
default setting for the MaltParser.
\item MSTParser, a second-order ``graph based'' (i.e., global score
  optimizing) parser \citep{mcdonald2005non,mcdonald2006online}.
\item Basic TurboParser \citep{martins2010turbo}, which is a first-order
  (arc-factored) model similar to the minimium
  spanning tree parser of \citet{mcdonald2005non}.
\item Standard TurboParser \citep{martins2011dual}, a second-order model that scores consecutive
  siblings and grandparents \citep{mcdonald2006online}. 
\item Full TurboParser \citep{martins2013turning}, which adds grand-sibling
  and tri-sibling (third-order)
  features as proposed by \citet{koo2010efficient} and implemented by
  \citet{martins2013turning}.
\item EasyFirst \citep{goldberg2010efficient}, a non-directional dependency parser
which builds a dependency tree by iteratively selecting
the best pair of neighbors to connect.\footnote{EasyFirst can only be trained
    to produce unlabeled dependencies.  It provides a labeler for SD
    version 1.6.5, but it cannot be retrained.  We therefore only
    report UAS for EasyFirst.}  
\item Huang's linear-time parser \citep{huang2010dynamic,huang2012structured}, a shift-reduce parser
that applies a polynomial-time dynamic programming algorithm that
achieves linear runtime in practice.\footnote{Huang's parser
    only produces unlabeled dependencies, so we only report UAS.}
\end{enumerate}
 POS tags for dependency parsers
were produced using version 2.0 of the Stanford POS Tagger (MEMM
tagging model ``left3words-wsj-0-18''; Toutanova et al.,
2003); \nocite{toutanova2003feature} this is identical to
\citet{cer2010parsing}.  POS tagging time and rules to transform
into \textsc{CCprocessed} graphs, where applied, are included in the
runtime.

Our comparison includes most of the parsers explored by
\citet{cer2010parsing}, and all of the top-performing ones. 
They found the Charniak-Johnson parser to be more than one
point ahead of the second best (Berkeley). MaltParser was the best
among d-parsing alternatives considered.

\begin{table*}
\begin{center}
{ \renewcommand{\arraystretch}{1.0}
\begin{tabular}{r|l|l|rrrr|r|}
\cline{2-8}
& \multirow{2}{*}{Type}   & \multirow{2}{*}{Parser}          & \multicolumn{2}{c}{PTB \S 22} & \multicolumn{2}{c|}{PTB \S 23} & Speed     \\  
  &                      &                                  & UAS         & LAS        & UAS          & LAS                & (tokens/s.) \\ \cline{2-8}
{\tiny 1} &\multirow{4}{*}{c-parsing} & Stanford englishPCFG         & $^\ddagger$90.06       & $^\ddagger$87.17      & $^\ddagger$90.09        & $^\ddagger$87.48                   & 123.63     \\  
 {\tiny 2} &                       & Stanford RNN                         & $^\ddagger$93.11       & $^\ddagger$90.16      & $^\ddagger$92.80        & $^\ddagger$91.10                  & 66.57    \\  
  {\tiny 3} &                       & Berkeley                         & 93.33       & 90.64      & 93.31        & 91.01                  & 200.00    \\  
           {\tiny 4} &             & Charniak-Johnson
           & $^\ddagger$ \bf 93.91       & $^\ddagger$ \bf 91.25      & $^\ddagger$ \bf 94.38        & $^\ddagger$ \bf 92.07                & 100.72    \\ \cline{2-8}
{\tiny 5} & \multirow{8}{*}{d-parsing}   & MaltParser (liblinear stackproj) & 88.93       & 86.23      & 88.47        & 85.90                  & 8799.91   \\ 
        {\tiny 6} &                & MaltParser (libsvm arc-eager)    & 89.35       & 86.61      & 89.20        & 86.46                & 8.81      \\ 
        {\tiny 7} &                & MSTParser                        & 91.24       & 87.83      & 90.87        & 87.53                & 239.60    \\ 
        {\tiny 8} &               & Basic TurboParser               & 90.25       & 87.93      & 90.12        & 87.89                 & 2419.98   \\ 
        {\tiny 9} &                & Standard TurboParser            & 92.16       & 89.50      & 91.95        & 89.40                & 413.67    \\ 
        {\tiny 10} &                & Full TurboParser
        &\bf 92.29       & \bf 89.64      & \bf 92.20        &  \bf 89.67                 & 209.98    \\ 
       {\tiny 11} &                 & EasyFirst                        & 89.80       & --          & 89.28        & --              & $^\ast$2616.19  \\ 
       {\tiny 12} &                 & Huang             & $^\dagger$90.90      & --          & $^\dagger$90.70       & --              & $^\ast$616.55   \\ \cline{2-8}
\end{tabular}}
\end{center}
\caption{\label{exp} \textsc{Basic} SD parsing performance 
  and runtime. 
  $^\ast$EasyFirst and Huang do not include dependency
  labeling in the runtime. $^\dagger$When training the Huang parser,
  we provide \S 23 (\S 22) as development data when testing on \S 22
  (\S 23), which gives a slight advantage. $^\ddagger$Charniak-Johnson uses semi-supervised training; Stanford englishPCFG and RNN models use external resources; see text.}
\end{table*}

\subsection{\textsc{Basic} Dependencies} \label{se:basic}

Table~\ref{exp} presents our results on \textsc{Basic} dependencies.
The most \emph{accurate} approach is still to use the
Charniak-Johnson parser (4), though Full TurboParser (10) is the best among
d-parsing techniques, lagging Charniak-Johnson by 2--3 absolute points
and with about twice the speed.
If the Stanford englishPCFG model 
provides adequate accuracy for a
downstream application, then we advise using MSTParser or any variant
of TurboParser
instead.  In particular, without sacrificing the Stanford englishPCFG's
level of performance, Basic TurboParser runs
nearly 20 times faster.

\begin{figure*}
\centering
\centerline{\includegraphics[width=5in]{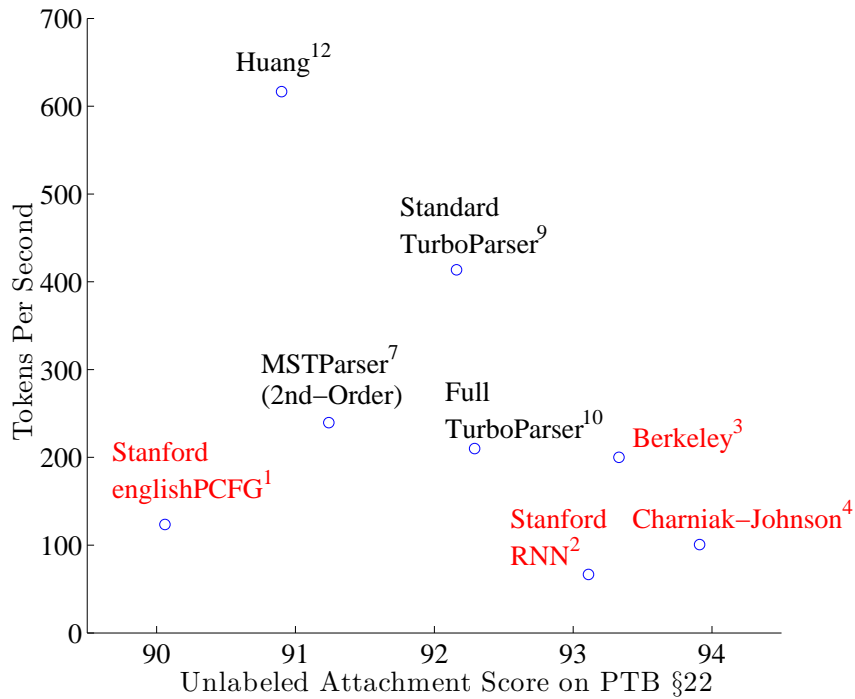}}
\caption{Speed-accuracy tradeoff. Huang, MSTParser, and all the three versions of
  TurboParser give better accuracy and speed than the ``default''
  Stanford CoreNLP pipeline. 
\label{fig:tradeoff}}
\end{figure*}

Figure~\ref{fig:tradeoff} plots the tradeoff between speed and
accuracy for most of the approaches.  For clarity, we exclude parsers
at the extremely fast and slow ends (all with accuracy around the same
or slightly below Stanford englishPCFG at the lower left of the plot).

\subsection{\textsc{CCprocessed} Dependencies} 
\label{se:ccprocessed}

\begin{table*}
\begin{center}
{ \renewcommand{\arraystretch}{1.0}
\begin{tabular}{r|l|l|rrrr|}
 \cline{2-7}
& \multirow{2}{*}{Type} & \multirow{2}{*}{Parser}  & \multicolumn{2}{c}{PTB \S 22} & \multicolumn{2}{c|}{PTB \S 23} \\  
  &          &                   & U.~$F_1$         & L.~$F_1$        & U.~$F_1$          & L.~$F_1$         \\ \cline{2-7}
{\tiny 1} &\multirow{3}{*}{c-parsing} & Stanford englishPCFG        & $^\dagger$87.7       & $^\dagger$84.6      & $^\dagger$87.9        & $^\dagger$85.0         \\  
  {\tiny 2} &                       & Stanford RNN                         & $^\dagger$91.3       & $^\dagger$88.1      & $^\dagger$91.0        & $^\dagger$88.1        \\  
 {\tiny 3} &                       & Berkeley                         & 91.4       & 88.5      & 91.4        & 88.9        \\  
   {\tiny 4} &             & Charniak-Johnson
           & $^\dagger$\bf 92.2       &$^\dagger$\bf 89.4      & $^\dagger$\bf 92.7        &$^\dagger$\bf 90.3      \\ \cline{2-7}
{\tiny 5} & \multirow{6}{*}{d-parsing}   & MaltParser (liblinear stackproj) & 86.3       & 83.2      & 85.4        & 82.5        \\ 
        {\tiny 6} &                & MaltParser (libsvm arc-eager)    & 86.2       & 83.2      & 86.0        & 83.2          \\ 
        {\tiny 7} &                & MSTParser                        & 87.1       & 83.5      & 87.3        & 83.8         \\ 
        {\tiny 8} &               & Basic TurboParser               & 87.0       & 84.1      & 86.7        & 84.0        \\ 
        {\tiny 9} &                & Standard TurboParser            & 90.2       & 87.3      & 89.8        & 87.0         \\ 
        {\tiny 10} &                & Full TurboParser
        &\bf 90.4       & \bf 87.4      & \bf 90.1        &  \bf 87.3         \\  \cline{2-7}
\end{tabular}}
\end{center}
\caption{\label{expccprocessed} \textsc{CCprocessed} SD parsing
  performance.  
    Since converting to \textsc{CCprocessed} SD requires labeled
  dependencies, EasyFirst and Huang do not join in the comparison
  here. $^\dagger$Charniak-Johnson uses semi-supervised training; Stanford englishPCFG and RNN models use external resources; see text.}
\end{table*}

Table~\ref{expccprocessed} shows performance scores 
for \textsc{CCprocessed} dependencies extracted from the parses in
\S\ref{se:basic}.  Because the number of attachments in a parser's
\textsc{CCprocessed} output may not equal the number in the
gold-standard tree, we follow the convention of reporting $F_1$
scores (unlabeled and labeled).
The additional runtime for this transformation is negligible, so we do
not report runtimes. The EasyFirst and Huang parsers cannot be
evaluated this way, since labeled \textsc{Basic} dependencies are
required for the transformation to \textsc{CCprocessed}.
The pattern is quite similar to the \textsc{Basic} SD experiment, with
the same top performers among c- and d-parsers. The
gap between c-parsing and d-parsing is 2.6\% unlabeled $F_1$ (3.0\% labeled).

\section{Part-of-Speech Effects} \label{se:pos}

We next consider the effect of POS tagging quality on
SD parsing
performance.   We focus on the Berkeley parser, which performed
strongly among c-parsing techniques and is amenable to substituting
its default POS tagger,\footnote{We explored Berkeley POS tags rather
than Charniak-Johnson because the Charniak-Johnson parser alters the
Penn Treebank POS tag set slightly.  (For example, it introduces tags AUX and
AUXG.)  A fair comparison would require
extra steps to control for this important difference.} and the two strongest
d-parsing models, Standard and Full TurboParser. 

First, we consider how these parsers perform with gold-standard POS
tags provided at test time.  Results are shown in the top three rows
of Table~\ref{poscomp}.  As expected, all methods perform better with
better POS tags.  More interestingly, the gap between the Berkeley
parser and Full TurboParser is essentially gone, with each showing a slight
lead on one of the two datasets.  

Next (second block in Table~\ref{poscomp}), we compared these three parsers, given the POS tags produced by
the \emph{Berkeley} parser.  Both TurboParsers gain about one point in each
score (compared to their performance with Stanford POS tags reported
earlier and repeated in the third block of Table~\ref{poscomp}) 
and generally match the performance of the Berkeley parser with its
own POS tags.

Further, we see that the Berkeley parser suffers a drop of
performance---about one point on each score---when provided
\emph{Stanford} POS tags (the same tags provided to TurboParser).
Given Stanford POS tags, the Berkeley parser and Full TurboParser
again perform about the same.

Taken together, these results suggest that future
work on improving part-of-speech representations (perhaps along the
lines of latent annotation methods already optimized for phrase
structure parsing in the Berkeley parser; Petrov
et al., 2006), \nocite{petrov2006learning} specifically for
Stanford dependency representations, might lead to further gains.
Further, joint inference between part-of-speech tags and d-parsing
might also offer improvements \citep{hatori2011incremental,li2011joint}.

\begin{table*}
\begin{center} 
{ \renewcommand{\arraystretch}{1.0}
\begin{tabular}{r|l|l|rrrr|rrrr|}
\cline{2-11}
&\multirow{3}{*}{POS Tags} & \multirow{3}{*}{Parser}  
& \multicolumn{4}{c|}{\textsc{Basic}} &
\multicolumn{4}{c|}{\textsc{CCprocessed}} \\
&&& \multicolumn{2}{c}{PTB \S 22} &
\multicolumn{2}{c|}{PTB \S 23} & \multicolumn{2}{c}{PTB \S 22} &
\multicolumn{2}{c|}{PTB \S 23} \tn
& &  & UAS & LAS & UAS & LAS  & U.~$F_1$ & L.~$F_1$ & U.~$F_1$  & L.~$F_1$\tn
\cline{2-11}
&\multirow{3}{*}{Gold} & Berkeley & 93.61 & 91.85 & \bf 93.65 & \bf
92.05 & 91.7 & 89.7 & \bf 91.8 & \bf 90.0\tn
&& Standard TurboParser & 93.55 & 91.94 & 93.31 & 91.73 & 91.7 & 89.8 & 91.4 & 89.5\tn
&& Full TurboParser  &\bf 93.79 &\bf  92.21 & 93.56 & 91.99 & \bf 92.0 & \bf 90.1 & 91.6 & 89.8\tn

\cline{2-11}
{\tiny 3} & \multirow{3}{*}{Berkeley} & Berkeley & 93.33       & 90.64      & \bf 93.31        & \bf 91.01
& 91.4 & 88.5 & \bf 91.4 & \bf 88.9
\tn &  & Standard TurboParser & 93.25       & 90.67      & 92.74        & 90.42 & 91.4 & 88.6 & 90.8 & 88.2
\tn  &  & Full TurboParser & \bf 93.46       & \bf 90.88      & 93.01        & 90.70 & \bf 91.6 & \bf 88.8 & 91.1 & 88.5
\tn 

\cline{2-11}
&\multirow{3}{*}{Stanford} & Berkeley & 92.26 & 89.46 & \bf 92.41 & \bf 89.87 & 90.3 & 87.3 & \bf 90.3 & \bf 87.6\tn
{\tiny 9} && Standard TurboParser & 92.16 & 89.50 & 91.95 & 89.40 & 90.2 & 87.3 & 89.8 & 87.0\tn
{\tiny 10} && Full TurboParser  &\bf  92.29 & \bf 89.64 & 92.20 & 89.67 & \bf 90.4 & \bf 87.4 & 90.1 & 87.3\tn 
\cline{2-11}
\end{tabular} }
\end{center}
\caption{\label{poscomp} Effects of POS on  SD parsing performance.
  Numerals in the leftmost column refer to rows in Table~\ref{exp}.}
\end{table*}

\section{Yamada-Matsumoto Features} \label{se:stacking}

As noted in \S\ref{se:intro}, dependency parsing algorithms have
generally been successful for YM parsing, which emphasizes syntactic
(and typically more local) relationships over semantic ones.  Given
that dependency parsing can be at least twice as fast as
phrase-structure parsing, we consider 
exploiting YM dependencies within a SD parser.  Simply put,
a YM dependency parse might serve as a cheap substitute for a
phrase-structure parse, if we can transform YM trees into SD trees.

Fortunately, the featurized, discriminative modeling families
typically used in dependency parsing are ready consumers of new
features.  The idea of using a parse tree produced by one parser to
generate features for a second was explored by
\citet{NivreMcDonald2008acl} and \citet{martins2008stacking}, and
found effective.  The technical approach is called ``stacking,'' and
has typically been found most effective when two different parsing
models are applied in the two rounds.  Martins et al.~released a
package for stacking with MSTParser as the second
parser,\footnote{\url{http://www.ark.cs.cmu.edu/MSTParserStacked}}
which we apply here.  
The descriptions of the second parser's
features derived from the first parser are listed in
Table~\ref{stackingfeatures}; these were reported by 
to be the best-performing on \S 22 in more extensive experiments
following from \citet{martins2008stacking}.\footnote{Personal communication.}

\begin{table*}[t]
\begin{center}
    \begin{tabular}{|l|p{4in}|}
    \hline
    Name         & Description                                           \\
    \hline 
PredEdge &   Indicates whether the candidate 
                                   edge was present, and what was 
                                    its label.  \tn\hline
Sibling & Lemma, POS, link label, distance, 
                                   and direction of attachment of the 
                                   previous and next predicted  siblings.  \tn\hline
Grandparents & Lemma, POS, link label, distance, 
                                   and direction of attachment of the 
                                   previous and next predicted  siblings.  \tn\hline
PredHead   & Predicted head of the candidate 
                                   modifier (if PredEdge = 0).                      \tn\hline
AllChildren & Sequence of POS and link labels 
                                   of all the predicted children of
                                   the candidate
                                   head.                            \tn
    \hline

    \end{tabular}
    \caption{\label{stackingfeatures}Features derived from the first
      parser, used in the second, in stacking.}
\end{center}
\end{table*}

The method is as follows:
\begin{enumerate}
\item 
Sequentially partition the Penn Treebank \S 02--22 into three parts ($P_1$,
  $P_2$, and $P_3$).
\item Train three instances of the first parser $g_1$, $g_2$, $g_3$
  using $P_2 \cup P_3$,
$P_1 \cup P_3$, and $P_1 \cup P_2$, respectively. Then parse each
$P_i$ with $g_i$.  These predictions are used to generate features for the second 
parser, $h$; the partitioning ensures that $h$ is never trained on a
first-round parse from a ``cheating'' parser.  \label{step2}
\item Train the second parser $h$ on \S 02--21, including the
  predictions from Step~\ref{step2}.
\item Train the first parser $g$ on \S 02--21.
\item To parse the test set, apply $g$, then $h$.
\end{enumerate}

In our experiments, we consider four different first parsers:
MSTParser (second order, as before) and MaltParser (liblinear stackproj), each
targeting YM and SD dependencies ($2 \times 2$ combinations).  The
second parser is always MSTParser.  These parsers were chosen because
they are already integrated in to a publicly released implementation
of stacked parsing by \citet{martins2008stacking}. For reference, the performance of
MaltParser and MSTParser on YM dependencies, on PTB \S 22--23, tagged by the
Stanford POS Tagger are listed in Table~\ref{ymacc}. 

\begin{table*}[t]
\begin{center} 
{ \renewcommand{\arraystretch}{1.0}
\begin{tabular}{r|l|rrrr|}
\cline{2-6}
& \multirow{2}{*}{Parser} 
& \multicolumn{2}{c}{PTB \S 22} &
\multicolumn{2}{c|}{PTB \S 23}  \tn 
 & & UAS & LAS & UAS & LAS  \tn\cline{2-6}
                      & MaltPaser-YM
        & 89.60      & 85.80      & 89.37        & 85.95
            \tn
                        & MSTParser-YM
        & 92.17       & 88.36      & 91.97        & 88.46
            \tn
\cline{2-6}
\end{tabular}
\caption{\label{ymacc}Accuracies of MaltParser and MSTParser on
  YM dependencies.}
}\end{center}
\end{table*}

Stacking results are shown in Table~\ref{stackingcomp}.  First, we find that
all four combinations outperform MSTParser on its own.  The gains are
usually smallest when the same parser (MSTParser) and representation (SD) are used
at both levels.  Changing either the first parser's representation (to
YM) or algorithm (to MaltParser) gives higher performance, but varying
the representation is more important, with YM features giving a 1.5\%
absolute gain on LAS over MSTParser.  The runtime is roughly doubled;
this is what we would expect, since stacking involves running two
parsers in sequence.

These results suggest that in future work, Yamada-Matsumoto
representations (or approximations to them) should be incorporated
into the strongest d-parsers, and that other informative intermediate
representations may be worth seeking out.

\begin{table*}[t]
\begin{center} 
{ \renewcommand{\arraystretch}{1.0}
\begin{tabular}{r|l|rrrr|r|}
\cline{2-7}
& \multirow{2}{*}{Parser} 
& \multicolumn{2}{c}{PTB \S 22} &
\multicolumn{2}{c|}{PTB \S 23} & Speed \tn 
 & & UAS & LAS & UAS & LAS & (tokens/s.) \tn\cline{2-7}
        {\tiny 5}                 & MaltPaser-SD
        & 88.93       & 86.23      & 88.47        & 85.90
        & 8799.91    \tn
        {\tiny 7}                 & MSTParser-SD
        & 91.24       & 87.83      & 90.87        & 87.53
        & 239.60    \tn
 & Stacked(MSTParser-SD, MSTParser-SD)  & 91.38 & 88.85 & 90.98 & 88.86 & 98.67 \tn
 & Stacked(MSTParser-YM, MSTParser-SD) & \bf 91.85 & \bf 89.40 & \bf 91.43 & \bf 89.10
 & 95.94 \tn 
 & Stacked(MaltParser-SD, MSTParser-SD) & 91.55 & 89.02 & 91.18 & 88.91 & 163.79 \tn
 & Stacked(MaltParser-YM, MSTParser-SD) & 91.61 & 89.09 & 91.13 & 88.83 & 164.07 \tn
\cline{2-7}
\end{tabular}
\caption{Integrating Yamada-Matsumoto dependencies into a
  \textsc{Basic} SD
  parser. Numerals in the leftmost column refer to rows in
  Table~\ref{exp}.  ``MaltParser'' refers to the liblinear stackproj version.
   \label{stackingcomp} 
}
}\end{center}
\end{table*}

\subsection{Incorporating Distributional Representations} \label{se:idr}

Distributional information has recently been established as a useful
aid in resolving some difficult parsing ambiguities.
In phrase-structure parsing, for example, \citet{socher2013parsing} improves the
Stanford parser by 3.8\% absolute $F_1$ score by injecting word vector
representations and compositional operations on them (captured by
recursive neural networks) into the parser's probabilistic context-free grammar.
In dependency parsing, \citet{koo2008simple} demonstrated that cluster features can
effectively improve the performance of dependency parsers. 

We
employed two types of Brown clustering \citep{brown1992class} features
suggested by Koo et al.:  4--6 bit cluster representations
used as replacements for POS tags and full bit strings used as
replacements for word forms.\footnote{The cluster strings we use are the same as used by
\citet{koo2008simple}; they are publicly available at
\url{http://people.csail.mit.edu/maestro/papers/bllip-clusters.gz}} We incorporated these features into
different variants of  TurboParser, including its second and third
order features.  Because these cluster representations are learned
from a large unannotated text corpus, the result is a semi-supervised
d-parser.

Table~\ref{semiresults} reports results 
on \textsc{Basic} SD parsing. Both Full TurboParser and Standard TurboParser get improvement
from the cluster-based features. We compare to the Stanford recursive
neural network parser.\footnote{We
use the most recent model (``englishRNN.ser.gz''), shipped with Stanford CoreNLP Package (v. 3.3.0).}
The Full TurboParser matches the performance of the Stanford RNN model with around
3 times the speed, and the Standard TurboParser is slightly behind the
Stanford RNN model but may provides
another reasonable accuracy/speed trade-off here.

Note that although both methods incorporating distributional representations, the methods and the unlabeled
corpora used to construct these representations are
different. \citet{socher2013parsing} uses the 25-dimensional vectors
provided by \citet{turian2010word} trained on a cleaned version of the RCV1 \citep{lewis2004rcv1} corpus with roughly 37
million words (58\% of the original size) using the algorithm of \citet{collobert2008unified}. 
\citet{koo2008simple} used the BLLIP corpus \citep{charniak2000bllip}, 
which contains roughly 43 million words of \emph{Wall Street Journal}
text with the sentences in the Penn Treebank removed.  These
differences imply that this comparison should be taken only as a
practical one, not a controlled experiment comparing the methods.

\begin{table*}[t]
\begin{center} 
{ \renewcommand{\arraystretch}{1.0}
\begin{tabular}{r|l|rrrr|r|}
\cline{2-7}
& \multirow{2}{*}{Parser} 
& \multicolumn{2}{c}{PTB \S 22} &
\multicolumn{2}{c|}{PTB \S 23} & Speed \tn 
 & & UAS & LAS & UAS & LAS & (tokens/s.) \tn\cline{2-7}
{\tiny 1} & Stanford basicPCFG         & 90.06       & 87.17      & 90.09        & 87.48              & 123.63     \\  
{\tiny 2} & Stanford recursive neural network
        & \bf 93.11       & 90.16      & \bf 92.80        & 90.10
        & 66.57    \tn \cline{2-7}
        {\tiny 9} &                 Standard TurboParser            & 92.16       & 89.50      & 91.95        & 89.40               & 413.67    \\ 

        & Standard TurboParser with Brown cluster features
        & 92.82       & 90.14      & 92.50        & 89.95
        & 243.83    \tn
        {\tiny 10} &                 Full TurboParser
        &\bf 92.29       &  89.64      & 92.20        &  89.67                  & 209.98    \\ 
        & Full TurboParser with Brown cluster features
        & 92.96       & \bf 90.31      & 92.75        & \bf 90.20
        & 179.26    \tn

\cline{2-7}
\end{tabular}
\caption{Incorporating distributional word representations into \textsc{Basic} SD
  parsing.    Numerals in the leftmost column refer to rows in
  Table~\ref{exp}. \label{semiresults}}
}\end{center}
\end{table*}

\section{Conclusion}

We conducted an extensive empirical comparison of different methods
for obtaining Stanford typed dependencies.  While the most accurate
method still requires phrase-structure parsing, we found that
developments in dependency parsing have led to a much smaller gap
between the best phrase-structure parsing (c-parsing) methods and the best direct
dependency parsing (d-parsing) methods.  Further experiments show that
part-of-speech tagging, which in the strongest phrase-structure
parsers is carried out jointly with parsing, has a notable effect on
this gap.  This points the way forward toward targeted part-of-speech
representations for dependencies, and improved joint
part-of-speech/dependency analysis.  We also found benefit from using
an alternative, more syntax-focused dependency representation
\citep{yamada2003statistical}, to provide features for Stanford
dependency parsing.  Overall, we find that direct dependency parsing can
achieve similar results to the Stanford CoreNLP pipeline (including
the new recursive neural network-based Stanford parser) at much
greater speeds.  The TurboParser models trained in this work are available for download at:  \url{http://www.ark.cs.cmu.edu/TBSD}

\section*{Acknowledgments}

This research was partially supported by NSF grants IIS-1352440 and
CAREER IIS-1054319.  The authors thank several anonymous reviewers, Andr\'e Martins,
and Nathan Schneider for
helpful feedback on drafts of the paper.

\bibliographystyle{plainnat}
\bibliography{acl2014.bib}

\begin{thebibliography}{46}
\providecommand{\natexlab}[1]{#1}
\providecommand{\url}[1]{\texttt{#1}}
\expandafter\ifx\csname urlstyle\endcsname\relax
  \providecommand{\doi}[1]{doi: #1}\else
  \providecommand{\doi}{doi: \begingroup \urlstyle{rm}\Url}\fi

\bibitem[Bj\"{o}rne et~al.(2009)Bj\"{o}rne, Heimonen, Ginera, Airola, Pahkkala,
  and Salakoski]{bjorne2009extracting}
J.~Bj\"{o}rne, J.~Heimonen, F.~Ginera, A.~Airola, T.~Pahkkala, and
  T.~Salakoski.
\newblock Extracting complex biological events with rich graph-based feature
  sets.
\newblock In \emph{Proc. of BioNLP the Workshop on Current Trends in Biomedical
  Natural Language Processing: Shared Task}, 2009.

\bibitem[Bohnet(2010)]{bohnet2010very}
B.~Bohnet.
\newblock Very high accuracy and fast dependency parsing is not a
  contradiction.
\newblock In \emph{Proc. of COLING}, 2010.

\bibitem[Brown et~al.(1992)Brown, Desouza, Mercer, Pietra, and
  Lai]{brown1992class}
P.~Brown, P.~Desouza, R.~Mercer, V.~Pietra, and J.~Lai.
\newblock Class-based n-gram models of natural language.
\newblock \emph{Computational linguistics}, 18\penalty0 (4):\penalty0 467--479,
  1992.

\bibitem[Cer et~al.(2010)Cer, de~Marneffe, Jurafsky, and
  Manning]{cer2010parsing}
D.~Cer, M.~C. de~Marneffe, D.~Jurafsky, and C.~D Manning.
\newblock Parsing to {S}tanford dependencies: Trade-offs between speed and
  accuracy.
\newblock In \emph{Proc. of LREC}, 2010.

\bibitem[{\c{C}etino\u{g}lu} et~al.(2010){\c{C}etino\u{g}lu}, Foster, Nivre,
  Hogan, Cahill, and Genabith]{cetinoglu2010lfg}
{\"{O}.}~{\c{C}etino\u{g}lu}, J.~Foster, J.~Nivre, D.~Hogan, A.~Cahill, and
  J.~Genabith.
\newblock {LFG} without {C}-structures.
\newblock In \emph{Proc. of TLT}, 2010.

\bibitem[Chang and Lin(2011)]{chang2011libsvm}
C.~Chang and C.~Lin.
\newblock Libsvm: a library for support vector machines.
\newblock \emph{ACM Transactions on Intelligent Systems and Technology},
  2\penalty0 (3):\penalty0 27, 2011.

\bibitem[Charniak and Johnson(2005)]{charniak2005coarse}
E.~Charniak and M.~Johnson.
\newblock Coarse-to-fine n-best parsing and maxent discriminative reranking.
\newblock In \emph{Proc. of ACL}, 2005.

\bibitem[Charniak et~al.(2000)Charniak, Blaheta, Ge, Hall, Hale, and
  Johnson]{charniak2000bllip}
E.~Charniak, D.~Blaheta, N.~Ge, K.~Hall, J.~Hale, and M.~Johnson.
\newblock Bllip 1987-89 wsj corpus release 1.
\newblock \emph{Linguistic Data Consortium, Philadelphia}, 2000.

\bibitem[Che et~al.(2012)Che, Spitkovsky, and Liu]{che2012comparison}
W.~Che, V.~I. Spitkovsky, and T.~Liu.
\newblock A comparison of {Chinese} parsers for {Stanford} dependencies.
\newblock In \emph{Proc. of ACL}, 2012.

\bibitem[Choi and McCallum(2013)]{choi2013transition}
J.~Choi and A.~McCallum.
\newblock Transition-based dependency parsing with selectional branching.
\newblock In \emph{Proc. of ACL}, 2013.

\bibitem[Collobert and Weston(2008)]{collobert2008unified}
R.~Collobert and J.~Weston.
\newblock A unified architecture for natural language processing: Deep neural
  networks with multitask learning.
\newblock In \emph{Proc. of ICML}, 2008.

\bibitem[de~Marneffe and Manning(2008)]{de2008stanford}
M.~C. de~Marneffe and C.~D Manning.
\newblock The {Stanford} typed dependencies representation.
\newblock In \emph{Proc. of COLING workshop on Cross-Framework and Cross-Domain
  Parser Evaluation}, 2008.

\bibitem[de~Marneffe et~al.(2006)de~Marneffe, MacCartney, and
  Manning]{de2006generating}
M.~C. de~Marneffe, B.~MacCartney, and C.~D. Manning.
\newblock Generating typed dependency parses from phrase structure parses.
\newblock In \emph{Proc. of LREC}, 2006.

\bibitem[Elming et~al.(2013)Elming, Johannsen, Klerke, Lapponi, Martinez, and
  S{\o}gaard]{elming2013down}
J.~Elming, A.~Johannsen, S.~Klerke, E.~Lapponi, H.~Martinez, and A.~S{\o}gaard.
\newblock Down-stream effects of tree-to-dependency conversions.
\newblock In \emph{Proc. of NAACL-HLT}, 2013.

\bibitem[Fan et~al.(2008)Fan, Chang, Hsieh, Wang, and Lin]{fan2008liblinear}
R.~Fan, K.~Chang, C.~Hsieh, X.~Wang, and C.~Lin.
\newblock Liblinear: A library for large linear classification.
\newblock \emph{The Journal of Machine Learning Research}, 9:\penalty0
  1871--1874, 2008.

\bibitem[Goldberg and Elhadad(2010)]{goldberg2010efficient}
Y.~Goldberg and M.~Elhadad.
\newblock An efficient algorithm for easy-first non-directional dependency
  parsing.
\newblock In \emph{Proc. of ACL}, 2010.

\bibitem[Graff(1995)]{graff1995north}
D.~Graff.
\newblock North american news text corpus, 1995.

\bibitem[Hatori et~al.(2011)Hatori, Matsuzaki, Miyao, and
  Tsujii]{hatori2011incremental}
J.~Hatori, T.~Matsuzaki, Y.~Miyao, and J.~Tsujii.
\newblock Incremental joint pos tagging and dependency parsing in chinese.
\newblock In \emph{Proc. of IJCNLP}, 2011.

\bibitem[Huang and Sagae(2010)]{huang2010dynamic}
L.~Huang and K.~Sagae.
\newblock Dynamic programming for linear-time incremental parsing.
\newblock In \emph{Proc. of ACL}, 2010.

\bibitem[Huang et~al.(2012)Huang, Fayong, and Guo]{huang2012structured}
L.~Huang, S.~Fayong, and Y.~Guo.
\newblock Structured perceptron with inexact search.
\newblock In \emph{Proc. of NAACL-HLT}, 2012.

\bibitem[Klein and Manning(2003)]{klein2003accurate}
D.~Klein and C.~D. Manning.
\newblock Accurate unlexicalized parsing.
\newblock In \emph{Proc. of ACL}, 2003.

\bibitem[Koo and Collins(2010)]{koo2010efficient}
T.~Koo and M.~Collins.
\newblock Efficient third-order dependency parsers.
\newblock In \emph{Proc. of ACL}, 2010.

\bibitem[Koo et~al.(2008)Koo, Carreras, and Collins]{koo2008simple}
T.~Koo, X.~Carreras, and M.~Collins.
\newblock Simple semi-supervised dependency parsing.
\newblock In \emph{Proc. of ACL}, 2008.

\bibitem[Lewis et~al.(2004)Lewis, Yang, Rose, and Li]{lewis2004rcv1}
D.~Lewis, Y.~Yang, T.~Rose, and F.~Li.
\newblock Rcv1: A new benchmark collection for text categorization research.
\newblock \emph{The Journal of Machine Learning Research}, 5:\penalty0
  361--397, 2004.

\bibitem[Li et~al.(2011)Li, Zhang, Che, Liu, Chen, and Li]{li2011joint}
Z.~Li, M.~Zhang, W.~Che, T.~Liu, W.~Chen, and H.~Li.
\newblock Joint models for chinese pos tagging and dependency parsing.
\newblock In \emph{Proc. of EMNLP}, 2011.

\bibitem[Martins et~al.(2008)Martins, Das, Smith, and
  Xing]{martins2008stacking}
A.~F.~T Martins, D.~Das, N.~A. Smith, and E.~P. Xing.
\newblock Stacking dependency parsers.
\newblock In \emph{Proc. of EMNLP}, 2008.

\bibitem[Martins et~al.(2010)Martins, Smith, Xing, Figueiredo, and
  Aguiar]{martins2010turbo}
A.~F.~T. Martins, N.~A. Smith, E.~P. Xing, M.~A.~T. Figueiredo, and P.~M.~Q.
  Aguiar.
\newblock Turbo parsers: Dependency parsing by approximate variational
  inference.
\newblock In \emph{Proc. of EMNLP}, 2010.

\bibitem[Martins et~al.(2011)Martins, Smith, Aguiar, and
  Figueiredo]{martins2011dual}
A.~F.~T. Martins, N.~A. Smith, P.~M.~Q. Aguiar, and M.~A.~T. Figueiredo.
\newblock Dual decomposition with many overlapping components.
\newblock In \emph{Proc. of EMNLP}, 2011.

\bibitem[Martins et~al.(2013)Martins, Almeida, and Smith]{martins2013turning}
A.~F.~T. Martins, M.~Almeida, and N.~A. Smith.
\newblock Turning on the turbo: Fast third-order non-projective turbo parsers.
\newblock In \emph{Proc. of ACL}, 2013.

\bibitem[McClosky et~al.(2006)McClosky, Charniak, and
  Johnson]{mcclosky2006effective}
D.~McClosky, E.~Charniak, and M.~Johnson.
\newblock Effective self-training for parsing.
\newblock In \emph{Proc. of NAACL-HLT}, 2006.

\bibitem[McDonald and Pereira(2006)]{mcdonald2006online}
R.~McDonald and F.~Pereira.
\newblock Online learning of approximate dependency parsing algorithms.
\newblock In \emph{Proc. of EACL}, 2006.

\bibitem[McDonald et~al.(2005)McDonald, Pereira, Ribarov, and
  Haji{\v{c}}]{mcdonald2005non}
R.~McDonald, F.~Pereira, K.~Ribarov, and J.~Haji{\v{c}}.
\newblock Non-projective dependency parsing using spanning tree algorithms.
\newblock In \emph{Proc. of HLT-EMNLP}, 2005.

\bibitem[Nivre(2004)]{nivre2004incrementality}
J.~Nivre.
\newblock Incrementality in deterministic dependency parsing.
\newblock In \emph{Proc. of ACL Workshop on Incremental Parsing: Bringing
  Engineering and Cognition Together}, 2004.

\bibitem[Nivre(2009)]{nivre2009non}
J.~Nivre.
\newblock Non-projective dependency parsing in expected linear time.
\newblock In \emph{Proc. of ACL-IJCNLP}, 2009.

\bibitem[Nivre and McDonald(2008)]{NivreMcDonald2008acl}
J.~Nivre and R.~McDonald.
\newblock Integrating graph-based and transition-based dependency parsers.
\newblock In \emph{Proc. of ACL-HLT}, 2008.

\bibitem[Nivre et~al.(2006)Nivre, Hall, and Nilsson]{nivre2006maltparser}
J.~Nivre, J.~Hall, and J.~Nilsson.
\newblock Maltparser: A data-driven parser-generator for dependency parsing.
\newblock In \emph{Proc. of LREC}, 2006.

\bibitem[Nivre et~al.(2009)Nivre, Kuhlmann, and Hall]{nivre2009improved}
J.~Nivre, M.~Kuhlmann, and J.~Hall.
\newblock An improved oracle for dependency parsing with online reordering.
\newblock In \emph{Proc. of IWPT}, 2009.

\bibitem[Petrov et~al.(2006)Petrov, Barrett, Thibaux, and
  Klein]{petrov2006learning}
S.~Petrov, L.~Barrett, R.~Thibaux, and D.~Klein.
\newblock Learning accurate, compact, and interpretable tree annotation.
\newblock In \emph{Proc. of COLING-ACL}, 2006.

\bibitem[Rush and Petrov(2012)]{rush2012vine}
A.~M. Rush and S.~Petrov.
\newblock Vine pruning for efficient multi-pass dependency parsing.
\newblock In \emph{Proc. of NAACL}, 2012.

\bibitem[Socher et~al.(2013)Socher, Bauer, Manning, and Ng]{socher2013parsing}
R.~Socher, J.~Bauer, C.~D. Manning, and A.~Ng.
\newblock Parsing with compositional vector grammars.
\newblock In \emph{Proc. of ACL}, 2013.

\bibitem[Stein et~al.(2010)Stein, Peitz, Vilar, and Ney]{stein2010cocktail}
D.~Stein, S.~Peitz, D.~Vilar, and H.~Ney.
\newblock A cocktail of deep syntactic features for hierarchical machine
  translation.
\newblock In \emph{Proc. of AMTA}, 2010.

\bibitem[Tetreault et~al.(2010)Tetreault, Foster, and
  Chodorow]{tetreault2010using}
J.~Tetreault, J.~Foster, and M.~Chodorow.
\newblock Using parse features for preposition selection and error detection.
\newblock In \emph{Proc. of ACL}, 2010.

\bibitem[Toutanova et~al.(2003)Toutanova, Klein, Manning, and
  Singer]{toutanova2003feature}
K.~Toutanova, D.~Klein, C.~D. Manning, and Y.~Singer.
\newblock Feature-rich part-of-speech tagging with a cyclic dependency network.
\newblock In \emph{Proc. of COLING-ACL}, 2003.

\bibitem[Turian et~al.(2010)Turian, Ratinov, and Bengio]{turian2010word}
J.~Turian, L.~Ratinov, and Y.~Bengio.
\newblock Word representations: a simple and general method for semi-supervised
  learning.
\newblock In \emph{Proc. of ACL}, 2010.

\bibitem[Yamada and Matsumoto(2003)]{yamada2003statistical}
H.~Yamada and Y.~Matsumoto.
\newblock Statistical dependency analysis with support vector machines.
\newblock In \emph{Proc. of IWPT}, 2003.

\bibitem[Zhang et~al.(2013)Zhang, Huang, Zhao, and McDonald]{zhang2013online}
H.~Zhang, L.~Huang, K.~Zhao, and R.~McDonald.
\newblock Online learning for inexact hypergraph search.
\newblock In \emph{Proc. of EMNLP}, 2013.

\end{thebibliography}

\appendix

\section{Additional Inference Rules} \label{app:rules}
We applied our additional inference rules after the collapsing transformation and
propagation of conjuncts transformation (implemented by the Stanford CoreNLP Toolkit).
The rules we use are as follows:
\begin{itemize}
\item{Rule 1:} Fixing the uncollapsed \emph{cc} dependencies: 
For each $\textit{cc}(A \rightarrow B)$, $T(A \rightarrow C)$, where $C$ is the first right sibling of $A$,
and $A$ linearly precedes $B$, add $\textit{conj\_\small{B}}(A \rightarrow B)$ and remove $T(A \rightarrow C)$.
\item{Rule 2:} Fixing the uncollapsed \emph{prep} dependencies:
For each $\textit{prep}(A \rightarrow B)$, $\textit{pobj}(B, C)$ and $A$ linearly precedes $B$ linearly precedes $C$,
add $\textit{prep\_\small{B}}(A \rightarrow C)$ and remove $\textit{prep}(A \rightarrow B)$, $\textit{pobj}(B \rightarrow C)$.
\end{itemize}

\begin{figure}
\centering
\centerline{\includegraphics[width=5in]{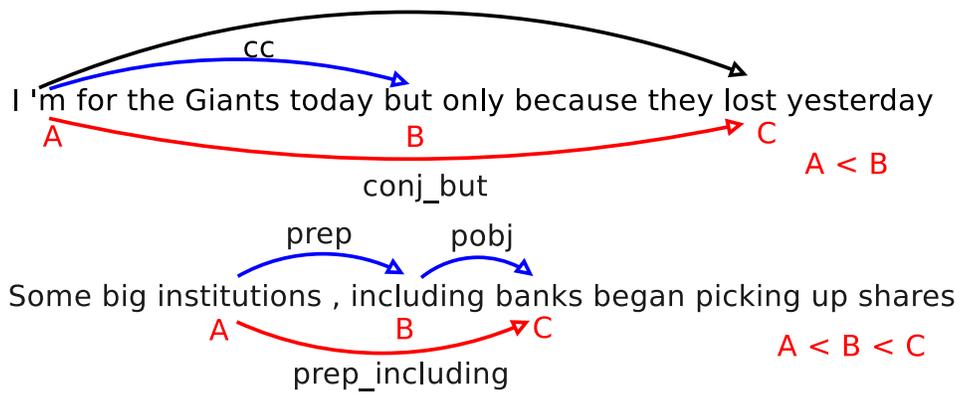}}
\caption{Examples of the application of rule 1 (above)
and 2 (below), showing the patterns above the sentence and the added dependencies
in {\color{red} red} below the text. Dependencies in {\color{blue} blue} will be removed. 
\label{fig:rules}}
\end{figure}

\end{document}